\documentclass{article}
\usepackage{spconf,amsmath,graphicx}

\usepackage{enumitem}
\usepackage{booktabs}
\usepackage{hyperref}
\usepackage[HTML]{xcolor}
\setlist{nosep, leftmargin=14pt}

\title{Unsupervised Detection of Post-Stroke Brain Abnormalities}
\name{Youwan Mah\'e$^{1,4}$ \quad Elise Bannier$^{1,3}$ \quad Stéphanie Leplaideur$^{1,2,6}$ \quad Elisa Fromont$^{5}$ \quad Francesca Galassi$^{1}$}

\address{%
$^{1}$ Univ Rennes, Inria, CNRS, Inserm, IRISA UMR 6074, Empenn, Rennes, France \\
$^{2}$ CHU Rennes, Physical Medicine and Rehabilitation Department, Rennes, France \\
$^{3}$ CHU Rennes, Radiology Department, Rennes, France \\
$^{4}$ Siemens Healthineers, Courbevoie, France \\
$^{5}$ Univ Rennes, Inria, CNRS, IRISA, Rennes, France \\
$^{6}$ Centre de Kerpape, Ploemeur, France \\
}

\begin{document}
%\ninept
%
\maketitle
\begin{abstract}
Post-stroke MRI not only delineates focal lesions but also reveals secondary structural changes, such as atrophy and ventricular enlargement. These abnormalities, increasingly recognised as imaging biomarkers of recovery and outcome, remain poorly captured by supervised segmentation methods. We evaluate REFLECT, a flow-based generative model, for unsupervised detection of both focal and non-lesional abnormalities in post-stroke patients. Using dual-expert central-slice annotations on ATLAS data, performance was assessed at the object level with Free-Response ROC analysis for anomaly maps. Two models were trained on lesion-free slices from stroke patients (ATLAS) and on healthy controls (IXI) to test the effect of training data. On ATLAS test subjects, the IXI-trained model achieved higher lesion segmentation (Dice = 0.37 vs 0.27) and improved sensitivity to non-lesional abnormalities (FROC = 0.62 vs 0.43). Training on fully healthy anatomy improves the modelling of normal variability, enabling broader and more reliable detection of structural abnormalities.

\end{abstract}
\begin{keywords}
Unsupervised Anomaly Detection, Magnetic Resonance Imaging, Stroke, Generative Modelling\end{keywords}
\section{Introduction}
\label{sec:intro}
Stroke remains a leading cause of long-term disability worldwide~\cite{Feigin2021}. In the sub-acute and chronic phases, morphological magnetic resonance imaging (MRI) is routinely used to assess brain damage and guide rehabilitation strategies~\cite{Bernhardt2017}. Morphological damage can extend beyond vascular lesions (infarcts) to include auxiliary changes such as white-matter hyperintensities (WMHs), cortical thinning, altered gyrification, and ventricular enlargement ~\cite{Tuladhar2015, Bonkhoff2022, Hu2021}. Although these secondary abnormalities are less well documented, they are increasingly recognised as relevant biomarkers for clinical assessment and recovery~\cite{Benali2023}.

Supervised deep-learning methods trained on large annotated datasets (e.g., ATLAS v2.0) have achieved state-of-the-art performance for infarct segmentation ~\cite{Liew2018,Dimatteo2025,Ahmed2023}. However, their focus remains on lesions, neglecting the broader spectrum of structural changes. In contrast, generative models trained on healthy anatomy can, in principle, flag any deviation from normal structure~\cite{Mahe2025}, enabling unsupervised exploration of diverse brain abnormalities. Nevertheless, to date, unsupervised anomaly detection approaches have almost exclusively been evaluated on focal lesions - a methodological gap that may partly explain why their Dice scores often lag behind supervised methods.

In this study, we evaluate the flow-based REFLECT model~\cite{Beizaee2025} for unsupervised detection of brain abnormalities in stroke patients beyond focal lesions. Using dual-expert annotations on ATLAS data, we assess both lesion segmentation and general anomaly detection, and examine how training data influence performance. Two variants were trained on lesion-free slices from stroke patients (ATLAS) and on healthy controls (IXI) to test whether learning from fully healthy anatomy improves sensitivity and generalisation across different types of brain abnormalities in stroke patients.

\section{Methods}
\label{sec:methods}
\subsection{Data and Annotations}
\label{subsec:dataset}
This study used two publicly available datasets: the ATLAS v2.0 dataset~\cite{Liew2018} and the IXI dataset\footnote{\url{http://brain-development.org/ixi-dataset/}}.
The ATLAS v2.0 dataset includes 655 T1-w MRIs (acquired at 1.5T and 3T across 33 imaging sites) from sub-acute and chronic stroke patients with manually annotated lesion masks. Following~\cite{Beizaee2025}, 80\% of the subjects were used for training, 10\% for validation, and 10\% for testing. The IXI dataset comprises 581 T1-w MRIs from healthy adults acquired at three London hospitals and was used to model healthy brain anatomy.
All MRIs from both datasets were resampled to 1\,mm isotropic resolution, registered to the MNI-152 template, corrected for intensity inhomogeneities using N4 bias correction, and z-score normalised. From each preprocessed volume, 20 central axial slices were extracted, skull-stripped, and zero-padded to 256$\times$256\,pixels.
For the ATLAS-trained model, all slices intersecting lesion masks were excluded during training to ensure that only apparently healthy tissue was used for learning. This filtering resulted in $3,384$ non-lesional ATLAS training slices compared with $11,620$ fully healthy IXI slices.
The validation set was used to estimate the optimal binarisation threshold, while the test set was used for evaluation.
Only the central slice of each subject from the ATLAS test subset (66 in total) was used for evaluation. 
We evaluated the central slice per subject to avoid lesion-centric bias and ensure consistent anatomical sampling, yielding a more challenging and representative evaluation. These slices were independently annotated by two experts (EB, MRI physicist; SL, rehabilitation physician) to mark both focal lesions and non-lesional abnormalities using point-based clicks, with a representative click placed at the most prominent location. The final reference set, obtained by merging the annotations from both raters (averaging the positions of annotations closer than 5 pixels), served as the ground truth for evaluation. Among the 66 slices, 62 contained at least one abnormality and were used for abnormality detection analysis, while 51 contained lesion masks and were used for lesion segmentation metrics. The remaining four normal slices were retained to estimate the baseline distribution of anomaly-map intensities for threshold calibration.

\subsection{Framework}
\label{subsec:framework}
We used \textbf{REFLECT}~\cite{Beizaee2025}, a state-of-the-art unsupervised brain anomaly segmentation framework based on rectified flows.  
Rectified flows~\cite{Xingchao2023}, extending continuous normalising flows~\cite{Chen2018}, learn an ordinary differential equation (ODE) that transports samples from an initial (pathological) distribution to a target (healthy) distribution along nearly straight trajectories in time.  
During inference, each pathological image is transported into the healthy distribution to generate a \textit{counterfactual reconstruction} - a pixelwise healthy version of the input. The anomaly map is then obtained as the difference between the input and its reconstruction.  
REFLECT is trained on non-pathological 2D axial slices using a self-supervised objective that matches the rectified flows between each slice and a synthetically lesioned counterpart.

\subsection{Training Protocol}
\label{subsec:training}
Model training followed the configuration in~\cite{Beizaee2025}, using the rectified-flow formulation for generative modelling. All networks were trained on 2D axial slices with a batch size of 30 (GPU-limited). We initialised the models with the official REFLECT VAE checkpoint (four scale factors) from the public repository, and performed inference with five reverse ODE correction steps.
Two models were trained for comparison. The first replicated the original REFLECT setup, using lesion-free slices from the ATLAS dataset. The second was trained on healthy MRIs from the IXI dataset to assess the role of training data composition. This setup enabled a direct comparison between models trained on lesion-free but stroke-affected brains and those trained on a fully healthy population.

\subsection{Evaluation}
\label{subsec:evaluation}
Model evaluation was performed in two complementary settings: (\textit{i}) lesion detection and segmentation, and (\textit{ii}) general anomaly detection in stroke patients.
Segmentation and detection use distinct thresholding strategies by design (task-specific operating points); thresholds are selected exclusively on validation/normal slices to avoid test leakage.

\subsubsection{Lesion detection and segmentation}
The optimal binarisation threshold for anomaly maps was determined on the validation set by maximising the Dice coefficient ($0.40$ and $0.47$ for the ATLAS- and IXI-trained models, respectively).  
Segmentation accuracy was evaluated using the Dice coefficient, 95th-percentile Hausdorff distance (HD95), and Average Surface Distance (ASD), providing complementary measures of overlap and boundary precision.  
Lesion-wise detection was assessed with the F1 score using a 10\% overlap criterion between predictions and reference masks.  
To analyse size-dependent performance, lesions were grouped into small (S), medium (M), and large (L) categories, defined by the 25th and 75th percentiles of the lesion size distribution~\cite{Bercea2025}.  
Statistical differences between models were tested using the paired Wilcoxon signed-rank test.

\subsubsection{Anomaly detection in stroke patients}
Detection performance was assessed using Free-Response Receiver Operating Characteristic (FROC) analysis, which measures sensitivity as a function of false positives per image (FPPI) and is suited for multiple anomalies per scan~\cite{Chakraborty2013}.  
Because expert annotations were point-based, evaluation was performed at the object level rather than by pixel overlap. A prediction was counted as a true positive if it contained, or lay within five pixels of, a reference annotation, allowing for minor localisation uncertainty. 
Anomaly maps were binarised at thresholds $T \in \{0.036, 0.1, 0.5\}$; the lowest threshold ($T=0.036$) was empirically derived from the mean plus three standard deviations of normal-slice intensities, approximating a 99\% confidence cutoff for normal variability.  
Each connected component was assigned a confidence score equal to its maximum pixel intensity.
Sensitivity was computed at FPPI levels of 0.25, 0.5, 1.0, and 1.5, and the final FROC score was defined as the mean sensitivity across these levels.

\section{Results}
\label{sec:results}
\subsection{Lesion segmentation and detection}
Performance results are summarised in Table~\ref{tab:performance_metrics}.
The IXI-trained model outperformed the ATLAS-trained model overall (Dice = $0.366$, F1 = $0.444$ vs $0.270$ / $0.326$), confirming that training on a larger and more diverse set of healthy images improves generalisation to unseen lesion patterns.
Performance increased with lesion size: large lesions (L) were reliably detected by both models (Dice ~ $0.60$), whereas small lesions (S) were frequently missed or only weakly detected, leading to very low Dice and F1 scores and inflated HD95/ASD values.
Medium-sized lesions (M) benefited most from IXI training, with significant gains in Dice ($+0.20$) and F1 ($+0.31$, $p<0.05$).
Boundary metrics (HD95 = $31.3$ mm, ASD = $10.9$ mm) were slightly higher for the IXI model, indicating marginally less precise contours compared with the ATLAS model (HD95 = $22.4$ mm, ASD = $4.18$ mm), although the difference was not statistically significant.

\begin{table}[ht]
    \centering
    \resizebox{\linewidth}{!}{
    \begin{tabular}{l c c c c}
    \hline
    & \textbf{Dice} & \textbf{HD95 (mm)} & \textbf{ASD (mm)} & \textbf{F1 10\%}  \\ \hline
    \multicolumn{5}{l}{\textbf{Trained on ATLAS}}\\
    \textbf{All} & 0.270 & \textbf{22.4} & \textbf{4.18} & 0.326 \\ 
    \textbf{S} & \textbf{0.074} & \textbf{5.64} & \textbf{2.67} & \textbf{0.153} \\ 
    \textbf{M} & 0.198 & \textbf{18.6} & \textbf{3.70} & 0.250\\ 
    \textbf{L} & 0.600 & \textbf{28.1} & \textbf{4.799} & \textbf{0.640} \\ \hline \hline
    \multicolumn{5}{l}{\textbf{Trained on IXI}}\\
    \textbf{All} & \textbf{0.366$^{\ast}$} & 31.3 & 10.9 & \textbf{0.444 }\\
    \textbf{S} & 0.060 & 54.8 & 40.8 & 0.077 \\
    \textbf{M} & \textbf{0.401$^{\ast}$} & 25.2 & 6.50 & \textbf{0.558$^{\ast}$} \\
    \textbf{L} & \textbf{0.609} & 32.6 & 7.4 & 0.603\\ \hline
    \end{tabular}}
    \caption{Lesion-wise segmentation (Dice, HD95, ASD) and detection (F1) metrics on ATLAS. \textbf{Bold} = best; $^{\ast}$ denotes $p<0.05$.}
    \label{tab:performance_metrics}
\end{table}
\subsection{Detection of brain abnormalities}
\begin{figure}[ht]
    \centering
    \includegraphics[width=\linewidth]{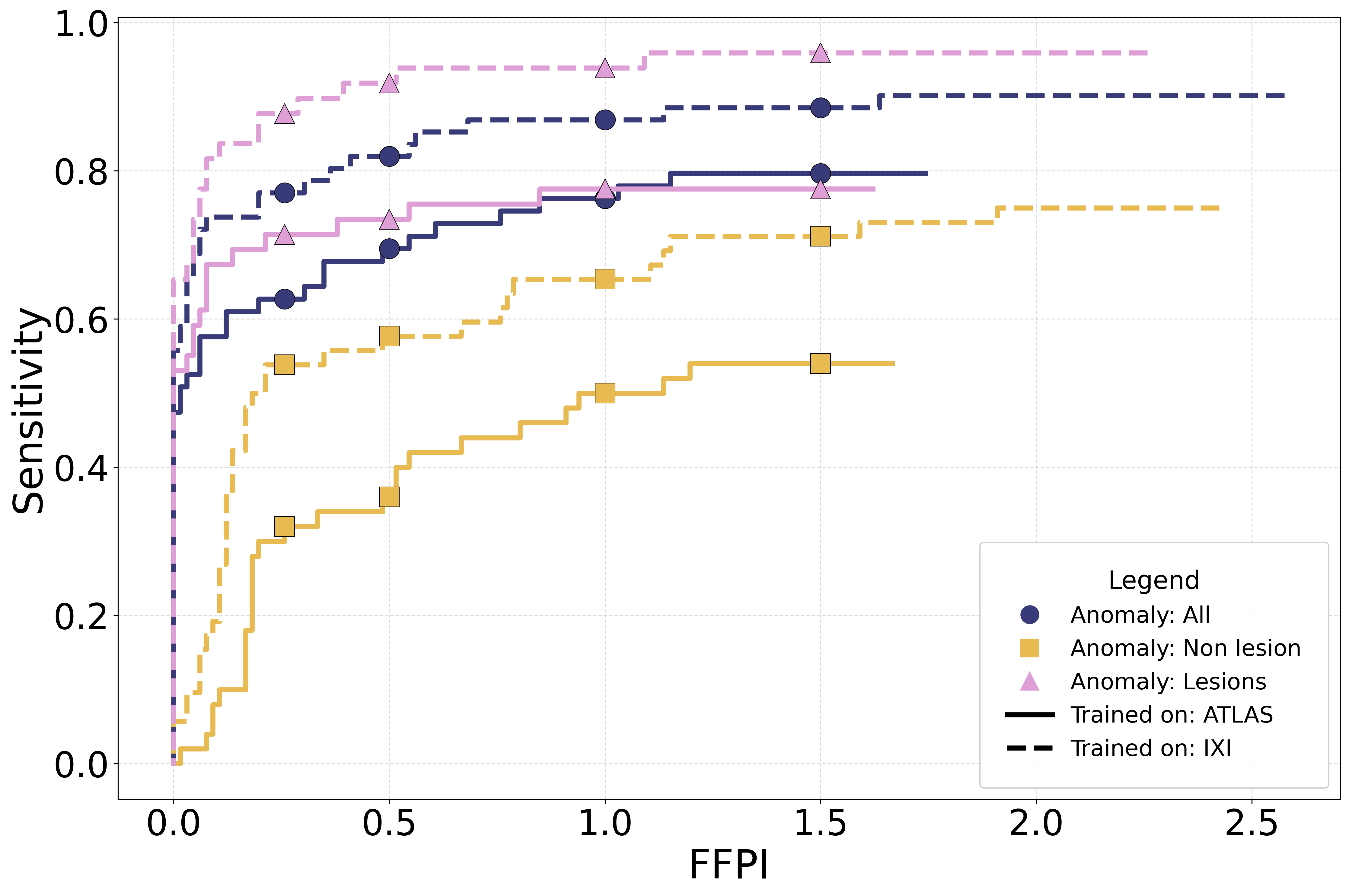}
    \caption{FROC curves comparing ATLAS- and IXI-trained models across FPPI levels 
    for overall, lesion and non-lesional abnormality detection at threshold \(T=0.036\).}
    \label{fig:FROC}
\end{figure}

\begin{figure}[ht]
    \centering
    \includegraphics[width=\linewidth]{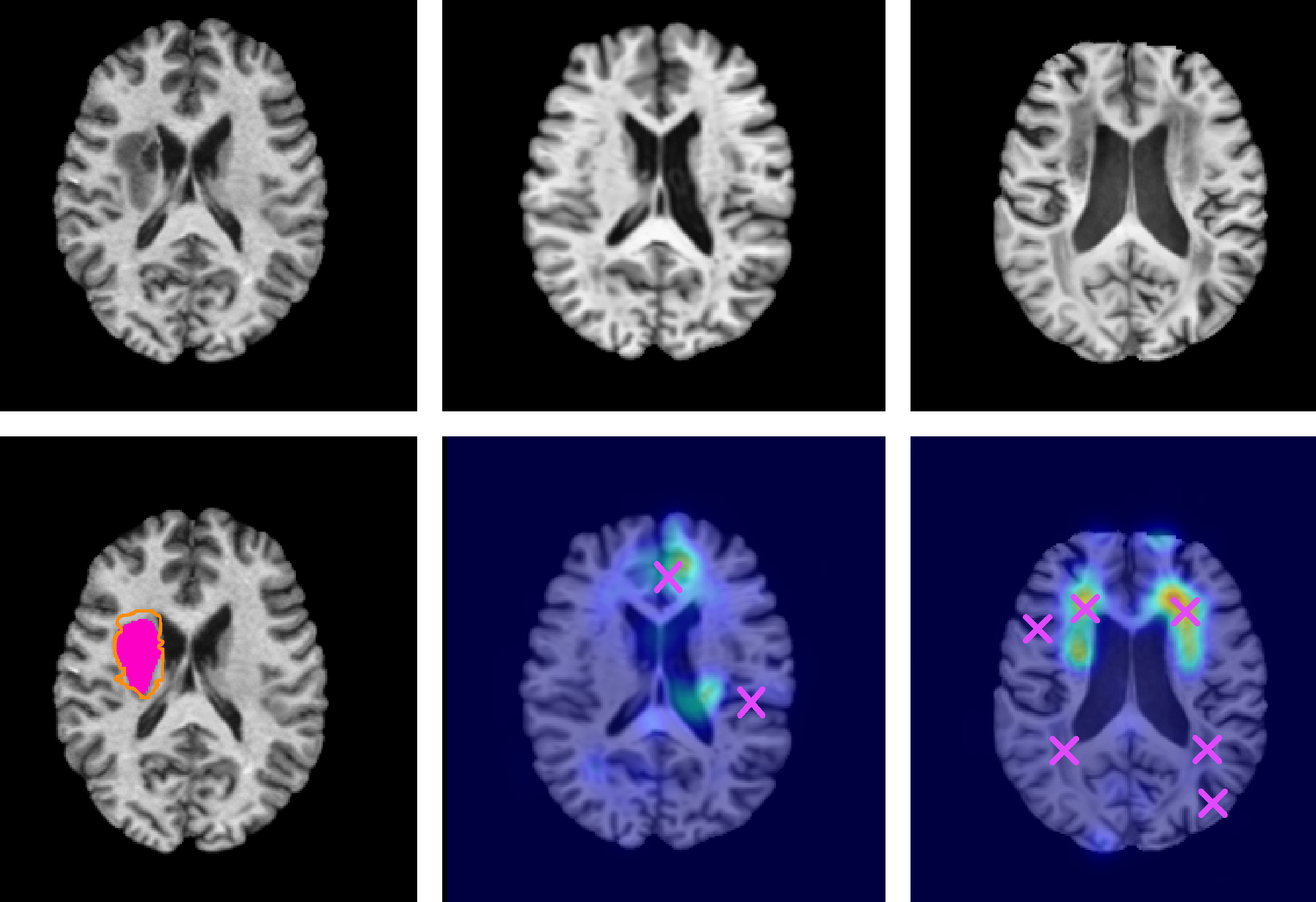}
    \caption{
    Examples of stroke-related anomaly detection from the ATLAS test set using the IXI-trained model. Top: pathological T1-w slices. Bottom, from left:~segmented lesion (Orange outline~=~Ground Truth, Pink~=~Prediction), anomaly heatmaps with expert annotations ($\times$), illustrating localisation of non-lesional abnormalities, including enlarged sulci (middle) and periventricular hyposignals (right).}
    \label{fig:AnoExample}
\end{figure}

FROC analysis (Fig.~\ref{fig:FROC}, Table~\ref{tab:FROC}) showed that both models accurately detected focal lesions, while the IXI-trained model achieved higher sensitivity to non-lesional abnormalities.
At the lowest threshold ($T=0.036$), the IXI model reached a FROC score of $0.620$ for non-lesional abnormalities versus $0.425$ for the ATLAS model. Sensitivity to non-lesional abnormalities decreased at higher thresholds, as expected for subtler signals. Lesion detection performance was highest at $T=0.5$, with FROC scores of $0.941$ (IXI) and $0.909$ (ATLAS). The IXI model maintained stable lesion detection across thresholds, whereas the ATLAS model showed larger variations.
These results indicate that training on a fully healthy population improves robustness and sensitivity to subtle abnormalities beyond focal lesions. Qualitative examples (Fig.~\ref{fig:AnoExample}) further illustrate accurate identification of both lesions and non-lesional regions.

\begin{table}[ht]
    \centering
    \resizebox{\linewidth}{!}{
    \begin{tabular}{lccc}
        \toprule
        \textbf{Method} & \textbf{$T=0.036$} & \textbf{$T=0.1$} & \textbf{$T=0.5$} \\
        \midrule
        \textbf{Trained on ATLAS} & & & \\
        Lesions only & 0.750 & 0.856 & 0.909 \\
        Non-lesion anomalies & 0.425 & 0.180 & 0.059 \\
        All anomalies & 0.720 & 0.794 & 0.909 \\
        \midrule\midrule
        \textbf{Trained on IXI} & & & \\
        Lesions only & 0.924 & 0.857 & \textbf{0.938} \\
        Non-lesion anomalies & \textbf{0.620} & 0.438 & 0.120 \\
        All anomalies & 0.836 & 0.791 & \textbf{0.941} \\
        \bottomrule
    \end{tabular}}
    \caption{FROC scores for anomaly detection across thresholds $T$. \textbf{Bold} = best.}
    \label{tab:FROC}
\end{table}

\section{Discussion}
\label{sec:discussion}
This study evaluated the REFLECT framework for unsupervised detection of focal and non-lesional abnormalities in stroke patients. Training on fully healthy IXI data improved sensitivity and generalisation compared with lesion-free ATLAS slices, highlighting the value of modelling a clean and diverse healthy distribution. Lesion-free stroke slices may still contain subtle pathology that biases the learned \textit{normal}. Part of this improvement may also reflect dataset scale, as IXI included $11,620$ healthy slices versus $3,384$ for ATLAS, offering a broader and more reliable representation of normal brain anatomy.
Both models accurately detected focal lesions, but sensitivity to diffuse or morphological abnormalities remained lower - consistent with training on synthetic, lesion-like perturbations. Lower binarisation thresholds were required to retain weak abnormality signals (maximum confidence ~$0.16$ for non-lesional regions vs. $0.64$ for lesions), reflecting a sensitivity–specificity trade-off typical of diffuse pathology detection.
Although Dice scores reached $0.6$ for large lesions, boundary distances (HD95 ~30 mm) remained high, largely due to small distant outliers rather than genuine contour errors. This effect is amplified by 2D evaluation, where isolated false positives on neighbouring slices inflate distance metrics despite good overall overlap. Dice scores were lower than those reported in the original REFLECT study ($0.41$), likely due to evaluation on central rather than lesion-focused slices.
Overall, these findings show that flow-based generative models can extend beyond lesion segmentation toward broader unsupervised detection of structural abnormalities in post-stroke patients. Future work should explore 3D implementations, adaptive thresholding, and multimodal MRI integration for improved clinical applicability.

\subsection{Limitations}
While promising, this study has several limitations. 
First, the use of 2D slices restricts spatial context, which may lead to misclassification in anatomically complex regions or for lesions spanning multiple planes. 
Second, the reliance on a fixed binarisation threshold introduces sensitivity to hyperparameter selection; optimal values may vary across datasets, scanners, and acquisition protocols due to domain shift. 
Third, evaluation relied on click-level rather than voxel-wise annotations, as precisely delineating diffuse abnormalities such as atrophy or ventricular enlargement is time-consuming and prone to inter-rater variability. This limitation may reduce validation precision and underestimate performance for subtle abnormalities. 
Future work should address these issues through 3D volumetric modelling, adaptive thresholding, and richer annotation frameworks to enable more robust and clinically deployable unsupervised abnormality detection.

\section{Conclusion}
\label{sec:conclusion}
In this work, we evaluated REFLECT, a flow-based generative model, for unsupervised brain abnormality detection and lesion segmentation in post-stroke MRI. 
Beyond confirming strong lesion detection performance, our results demonstrate that rectified-flow generative reconstruction models can also identify non-lesional abnormalities such as atrophy and ventricular enlargement - features often overlooked by conventional supervised, lesion-focused methods. Training on a healthy population dataset (IXI) improved generalisation and sensitivity across abnormality types, underscoring the importance of modelling normal anatomical variability.
Although sensitivity to sparse or global brain abnormalities remains limited compared to focal lesions, these findings highlight the potential of unsupervised reconstruction-based methods to provide complementary information for clinical assessment and recovery monitoring. Such unsupervised detection could support longitudinal monitoring of secondary degeneration and recovery, providing complementary information to standard lesion analyses. Future work should extend this framework toward full 3D volumetric modelling and incorporate anatomically informed or morphologically diverse synthetic perturbations, as well as hybrid training schemes integrating healthy and pathological data.

\section{Acknowledgements}
Experiments presented in this paper were carried out using the Grid’5000 testbed, supported by a scientific interest group hosted by Inria and including CNRS, RENATER and several universities as well as other organisations (see \url{https://www.grid5000.fr}).

\textbf{Compliance with Ethical Standards:} This study was conducted retrospectively using human MRI data publicly available through open-access datasets (ATLAS v2.0\footnote{\url{https://fcon_1000.projects.nitrc.org/indi/retro/atlas.html}} and IXI\footnote{\url{https://brain-development.org/ixi-dataset/}}). In accordance with the licensing terms of these datasets, separate ethical approval was not required.

\textbf{Conflicts of Interest:} The authors have no relevant financial or non-financial interests to disclose.
\bibliographystyle{IEEEbib}
\bibliography{refs}

\end{document}